\documentclass[runningheads]{llncs}

\usepackage{mathtools}
\usepackage{multirow}
\usepackage{graphicx}
\usepackage[tight,footnotesize]{subfigure}
\usepackage{epstopdf}
\usepackage{graphicx}
\usepackage[ruled,vlined]{algorithm2e}

\usepackage{svg}

\usepackage{xcolor}

\usepackage{enumitem}
\usepackage{url}

\usepackage{cleveref}
\Crefname{algorithm}{Alg.}{Algs.}
\Crefname{equation}{Eq.}{Eqs.}
\Crefname{section}{Sec.}{Secs.}
\Crefname{figure}{Fig.}{Figs.}
\Crefname{tabular}{Tab.}{Tabs.}
\Crefname{table}{Tab.}{Tabs.}

\begin{document}
\title{Keep your Distance: Determining Sampling and Distance Thresholds in Machine Learning Monitoring}
\titlerunning{Keep your Distance: A SafeML Framework}
\author{Al-Harith Farhad\inst{1}\orcidID{0000-0001-6749-0620 } \and
Ioannis Sorokos\inst{2}\orcidID{0000-0003-2704-8381} \and
Andreas Schmidt \inst{2}\orcidID{0000-0002-7113-7376} \and
Mohammed Naveed Akram\inst{2}\orcidID{0000-0002-0924-5536} \and 
Koorosh Aslansefat \inst{3}\orcidID{0000-0001-9318-8177} \and
Daniel Schneider\inst{2}\orcidID{0000-0003-3465-9738}
}
\authorrunning{H. Farhad et al.}
%
\institute{University of Mannheim, Schloss, 68131 Mannheim, Germany \email{afarhad@mail.uni-mannheim.de} \and
Fraunhofer IESE, Fraunhofer-Platz 1, 67663 Kaiserslautern, Germany \email{\{ioannis.sorokos, andreas.schmidt, naveed.akram, daniel.schneider\}@iese.fraunhofer.de} \and
University of Hull, Cottingham Rd., HU6 7RX Hull, UK \email{k.aslansefat@hull.ac.uk}
}
\maketitle              
\begin{abstract}
Machine Learning~(ML) has provided promising results in recent years across different applications and domains.
However, in many cases, qualities such as reliability or even safety need to be ensured.
To this end, one important aspect is to determine whether or not ML components are deployed in situations that are appropriate for their application scope.
For components whose environments are open and variable, for instance those found in autonomous vehicles, it is therefore important to monitor their operational situation to determine its distance from the ML components' trained scope.
If that distance is deemed too great, the application may choose to consider the ML component outcome unreliable and switch to alternatives, e.g. using human operator input instead.
SafeML is a model-agnostic approach for performing such monitoring, using distance measures based on statistical testing of the training and operational datasets.
Limitations in setting SafeML up properly include the lack of a systematic approach for determining, for a given application, how many operational samples are needed to yield reliable distance information as well as to determine an appropriate distance threshold.
In this work, we address these limitations by providing a practical approach and demonstrate its use in a well known traffic sign recognition problem, and on an example using the CARLA open-source automotive simulator.

\keywords{Machine Learning \and Monitoring \and Safety \and Uncertainty}
\end{abstract}


\section{Introduction} 
\label{sec:intro}

The continuous expansion of application fields of \emph{Machine Learning}~(ML) into safety-critical domains, such as autonomous vehicles, entails an increasing need for suitable safety assurance approaches. One key aspect in this regard is to get a grasp on the confidence associated with an output of a ML component.
While some ML models provide a probabilistic output that can be interpreted as a level of confidence, such output is not alone sufficient to establish overall trust.
Significant progress has been made towards addressing this question, with approaches that introduce more sophisticated evaluation of a given model's outputs.
Model-specific approaches base their evaluation on understanding of the internals of the given ML model, e.g.~\cite{paterson2021detection} focus on the second-to-last layer of a given deep neural network.
On the other hand, model-agnostic approaches treat models as black-boxes, basing their evaluation on properties that can be interrogated externally, e.g. in \cite{klas2019uncertainty}, surrogate models are constructed during training to later provide uncertainty estimates of the ML model in question.
An additional concern for evaluating ML models, is that the evaluation must also satisfy the application requirements, in particular with regards to performance.
For instance, the authors in  \cite{Rausch2021Autoencoder} propose auxiliary networks for evaluation, but the computational capacity needed to estimate them hinders their roll-out into real-time systems.
On a general note, the approaches discussed here, including ours, can be building blocks in assuring important quality properties of ML components and maybe even safety, but they will hardly be the silver bullet that solves the challenge alone.
A safety argument for a system with ML components will typically be very specific for a given application and its context and comprise of a diverse range of measures and assumptions, many of which we would expect to include both development-time approaches, as well as runtime approaches, with ours falling under the latter category. 

SafeML, proposed in \cite{Aslansefat2020safeml} and improved in \cite{aslansefat2021toward}, is a runtime approach for evaluating ML model outputs.
In brief, SafeML compares training and operational data of the ML model in question, and determines whether they are statistically 'too distant' to yield a trustworthy answer.
The work in \cite{aslansefat2021toward} further demonstrates a bootstrap-based p-value estimation extension to improve confidence in measurements.
However, the existing literature does not explain how specific challenges for practical application of SafeML are addressed.

Our contribution is to identify these limitations and propose an approach that enables a systematic application of SafeML and overcomes them.
For the remainder of \Cref{sec:intro}, we provide a more detailed description of previous work on SafeML.
We then discuss what its practical limitations are, provide the motivation behind our approach, and then further detail our contributions.

\subsection{SafeML} 
\label{sec:safeML}

SafeML is a collection of measures that estimate the statistical distance between training and operational datasets based on the \emph{Empirical Cumulative Distribution Function}~(ECDF).
In \cite{Aslansefat2020safeml}, the estimated distance has been shown to negatively correlate with a corresponding ML model’s accuracy.
In the same paper, a plausible workflow of applying SafeML for monitoring ML was also proposed. 
Per the workflow, an ML task can be divided into two phases, an offline/training phase and
an online/application phase.
In the training phase, it is assumed that we have a trusted dataset and there is no uncertainty associated with its labels.
An ML model, such as a deep neural network or a support vector machine, can be trained using the trusted data.
Assuming the model passes its validation testing, then it can be used in the online phase.
Also in the training phase, the ECDF of each feature and each class is stored for later comparison in the online/application phase.

In the online/application phase, the same trained model and a buffer is provided to gather a sufficient number of samples from inputs.
The number of buffered samples should be enough such that the distance determination can be relied upon, but the existing approach does not provide further guidance on how this number should be specified. 
When a large enough number of samples is obtained, based on the trained classifier decisions, the ECDF of each feature and each class is calculated.
The ECDF-based statistical distance measures are used to evaluate the differences between the trusted dataset and the buffered data.
To ensure that the statistical measures are valid, a bootstrap-based p-value evaluation method is added to the measurements, per~\cite{aslansefat2021toward}.
The user of the method must then specify a minimal distance threshold~(and additional ones optionally) for the distance measures.
The proposed workflow suggests that if the outcome is slightly above the minimal threshold, additional data can be requested.
Instead, if the outcome is significantly above the threshold value~(or a specified additional threshold), alternative actions can be taken, e.g. operator intervention.
If the outcome is below the minimal threshold~(or a specified additional threshold), the decision of the machine learning algorithm can be trusted and the statistical distance measures can be stored to be reported. 

SafeML, being model-agnostic, can be flexibly deployed in numerous applications.
In \cite{aslansefat2021toward,Aslansefat2020safeml}, Aslansefat et al. already presented experimental applications of SafeML on security attack detection~\cite{sharafaldin2018toward}, and \emph{German Traffic Sign Recognition Benchmark}~(GTSRB) examples \cite{stallkamp2012man}.
For security intrusion detection, SafeML measures are used to compare the statistical distances against the accuracy of classifier.
In the GTSRB example, the model is trained, and the incorrectly classified set of images are compared against randomly selected input images from the training set.

\subsection{Motivation}
\label{sec:motivation}

As mentioned in \Cref{sec:safeML}, applying SafeML requires specification of the number of runtime samples needed to be acquired, and at least the minimal distance threshold for acceptance/rejection.
Both parameters must be defined during development time, as they need to be known by the time the ML model is in operation.
Existing work in SafeML does not investigate nor provide guidance for establishing these parameters, leaving it up to the user to find reasonable values.

However, this is not a trivial matter, as identifying appropriate thresholds has application-related implications.
As will be highlighted further in \Cref{sec:methodology}, an inadequate number of runtime samples may results in low statistical power of the SafeML-based evaluation, whereas collecting too many samples can be inefficient and limit application performance. Addressing these limitations is the focus of this publication.

Statistical power is the probability of a correctly rejected null-hypothesis test i.e. the probability of a true positive, given a large enough population \cite{ellis_2010Statistical}.
Conversely, by presetting a required level of statistical power, the population size needed to correctly distinguish two distribution can be calculated through power analysis.
Similarly, distance thresholds that are too low can lead to overflowing the host application with false positive alarms, whereas distance thresholds that are too high can lead to overlooking potentially critical conditions.
Concretely, we establish the following research questions:
\begin{enumerate}[leftmargin=30pt]
    \item[\textbf{RQ1:}] \textbf{Dissimilarity-Accuracy Correlation}. Can we confirm that data points seen during operation that are dissimilar to training data impact the model's performance in terms of accuracy?
    \item[\textbf{RQ2:}] \textbf{Sample Size Dependency}. Can we determine whether the sample size affects the accuracy of the SafeML distance estimation?
\end{enumerate}

\subsection{Paper Contribution and Outline}
\label{sec:paperStructure}

The contribution of the paper is three-fold.
First, we use power analysis to specify sampling requirements for SafeML monitoring at runtime.
Secondly, we systematically determine appropriate SafeML distance thresholds.
Finally, we apply the above method in the context of an exemplary automotive simulation.

The remainder of the paper is as follows: In \Cref{sec:background}, we discuss background and related work, including approaches both similar, and differing to SafeML.
In \Cref{sec:methodology}, we describe our approach for systematically applying SafeML and determining relevant thresholds, as well as our experimental setup.
In \Cref{sec:results}, we discuss our experimental results before we recap our key points and discuss future work in \Cref{conclusion}.


\section{Background and Related Work} 
\label{sec:background}

To briefly recap, in \cite{aslansefat2021toward,Aslansefat2020safeml} the authors propose statistical distance measures to compare the distributions of the training and operational datasets; the measures are based on established two-sample statistical testing methods, including the Kolmogorov-Smirnov, Anderson-Darling, Cramer von Mises \cite{evans2017distribution}, and Wasserstein methods \cite{ramdas2017wasserstein}.
The statistical distance measures used by SafeML capture the dissimilarity between different two distributions, but the approach itself does not propose an explicit threshold at which those distributions are not equivalent, nor a means of how to determine one systematically.\\
Setting meaningful thresholds is a reoccurring problem in ML and data-driven applications.
A method based on the 3-sigma rule was shown to provide suitable threshold criteria in Hidden Markov Models under the assumption of normal distribution \cite{Duan2009Threshold}.
Our approach is similar in the sense that we used the same principle, but we did not assume that our datasets are normally distributed. Therefore, instead of a 3-sigma rule, we opted for a gradual increase of the threshold based on the sigma value. We elaborate on this further in \Cref{sec:methodology}.

A prerequisite for transition of AI applications into safety- and security-critical systems is having available guarantees and guidelines to assure underlying system dependability. A method was proposed in \cite{Rausch2021Autoencoder} to assure the model's operation within the intended context in a model-agnostic manner, where an additional autoencoder-based network is used to detect semantic novelty.\\
However, the innate problem of using neural networks, including autoencoders, is their black-box nature with respect to explainability, which inhibits the establishment of dependability guarantees. Hence, the use of a more explainable statistical method could serve as a solution to this issue. This includes our proposed approach here, as the ECDF-based distance to the training set could provide additional insight to the model's decision.

In \cite{paterson2021detection}, the authors propose a commonality metric, which inspects the second-to-last layer of a \emph{Deep Neural Network}~(DNN). The proposed metric expresses the ratio between the activation of the neurons in the last layer during training (across all training instances), versus their activation during operation, for the given operational input. The approach shares common ideas with SafeML, but diverges in terms of being model-specific, as the metric samples directly the last layer's neurons. 
Instead, SafeML does not consider the model internals, and makes no assumption on the distribution of the training and operational data.

Efforts have been made to ensure a dependable and consistent behavior in AI-based applications. These took various forms, from providing generative models, whose outputs can be interpreted as a confidence of predictions to the aforementioned novelty detection. However, design-time safety measures have been introduced in \cite{singh2019certifyNN}, where the robustness of neural networks could be certified through a novel abstract domain, before deployment. 
Similarly, a feature-guided safety testing method of neural network was proposed in \cite{wicker2018feature} to evaluate the robustness of neural network by feeding them through adversarial examples. Markov decision processes were also proposed to be paired with neural networks to verify their robustness through statistical model checking \cite{Gros2020Modelchecking}.

Uncertainty wrappers are another notable concept \cite{jockel2019increasing,jockel2019safe,klas2020framework,klas2019uncertainty}. This mathematical concept distinguishes ML uncertainty into three layers I) model performance, II) input quality and III) scope compliance, and provides a set of useful functions to evaluate the existing uncertainties in each step. The uncertainty wrapper can be compared with SafeML in the third layer (scope compliance). Both of them are model-agnostic. 

Safeguard AI \cite{lee2018training} proposes to calculate the likelihood of \emph{out-of-distribution} ~(OOD) inputs, and add it to the loss function of the ML/DL model.
This approach also uses a \emph{Generative Adversarial Network}~(GAN) to produce boundary data in order to create a more accurate OOD.
In comparison to SafeML, the approach is model-specific and cannot be evaluated at runtime.

Another common theme across approaches for safeguarding ML models is investigating all conceivable input perturbations and produce robust, safe, and abstract interpretable solutions and certifications for ML/DL models~\cite{gehr2018ai2,ruoss2020efficient,fischer2019dl2,mirman2019provable,mirman2018differentiable,muller2021prima}.
These approaches are also model-specific and do not provide runtime solutions.
Similar to previous approaches, \emph{DeepImportance} is a model-specific solution that presents a new \emph{Importance-Driven Criteria}~(IDC) as a layer-wise function to be assessed during the test procedure and provided a systematic framework for ML testing~\cite{gerasimou2020importance}.
Regarding the reliability evaluation of ML models, only a small number of solutions are provided so far and one them is ReAsDL.
ReAsDL divides the input space into tiny cells and evaluates the ML/DL reliability based on the cells' robustness and operational profile probability~\cite{zhao2021assessing,zhao2021detecting}.
This solution is model-agnostic and has focused on classification tasks similar to SafeML.
The NN-Dependability-kit has suggested a new set of dependability measures to assess the impact of uncertainty reduction in the ML/DL life cycle.
They also include a formal reasoning engine to ensure that the ML/DL dependability is guaranteed.
The approach can be used for runtime purposes~\cite{cheng2019nn}. 


\section{Methodology}
\label{sec:methodology}

In this section, we present our refined approach for applying SafeML, in the form of a proposed workflow, and address the question of determining the sampling and distance thresholds. 
To validate our approach, we applied SafeML both towards ML monitoring during simulation, as well as using it against an existing dataset, the GTSRB.
In the next section, we describe the experimental design for our empirical evaluation of the proposed approach.

\subsection{Process Workflow} 
\label{sec:flowchart}

\begin{figure}[t]
\includegraphics[width=\textwidth]{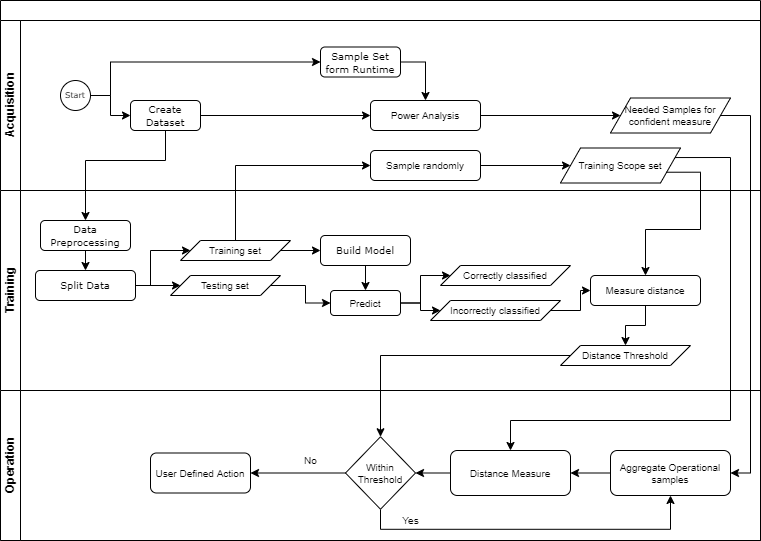}
\caption{Process flowchart} \label{fig:workflow}
\end{figure}

The process workflow to determining the needed number of samples as well as the distance threshold is divided into three stages as shown in \Cref{fig:workflow}.

\begin{itemize}
    \item \textbf{Acquisition:} In this stage, two datasets are involved, a training dataset and a testing dataset. In our empirical experiments (see \cref{sec:setup}), these datasets are generated from the simulation, but in general they should be derived during development. At this point, power analysis is used to find the number of samples to determine the difference between the operational and training set. 
    In general, this factor can be calibrated for the application at hand, as it determines an additional number of samples beyond the minimum needed to achieve the determined test power. The effect size for the power analysis is established between the training and testing set, using Cohen's d coefficient \cite{cohen1992power}.
    \item \textbf{Training:} The training dataset is processed and split into a training set and a testing set. A sub-sample of the smaller training set is uniformly sampled to represent the \emph{Training Scope Set}~(TSS) in the calculation of statistical distances to reduce computational complexity during runtime. A model is then built from the smaller training set and used to predict the outputs of the testing set. The result is further distinguished into correctly and incorrectly classified outputs, where SafeML measures evaluate the statistical distance between the incorrectly classified outputs and the TSS. The resulting distances are finally used as the initial distance threshold.
    \item \textbf{Operation:} Once the trained model is in operation, the value obtained in the Acquisition stage is used to aggregate operational data points into an operational set. SafeML measures evaluate the statistical distance between this operational set and the TSS. If the value falls within the defined threshold, the model continues its operation normally, otherwise, a signal is sent to run a user-defined action.
\end{itemize}

\subsection{Experiment Setup} \label{sec:setup}

We performed experiments on German Traffic Sign Recognition Benchmark (GTSRB)\cite{Stallkamp2012} and on a synthetic example dataset in the CARLA simulator\footnote{\url{https://carla.org}} \cite{Dosovitskiy17} for evaluating our approach.
CARLA is an automotive simulator used for the development, training, and validation of autonomous driving systems. The dataset generated from CARLA was used to evaluate the confidence level of SafeML predictions and the autopilot decisions of the simulated vehicle. The GTSRB dataset is a collection of traffic sign images, along with their labels used for benchmarking the ML algorithms. It was first used in 2011. The dataset is a good representation of safety-critical application of ML-components. Hence, it was also considered in this work for the evaluation of the approach presented. 

The CARLA setup allows us to identify a systematic method for estimating the minimum number of required samples and the distance acceptance threshold though a fixed-point iteration as well as their implication on the model's prediction, and how they correlate to the model's performance.
A simple model was built from a dataset sampled from CARLA, using a vehicle autopilot with varying driver profiles (shown in \Cref{driverProf}). This corresponds to the `Acquisition' step in section \cref{sec:flowchart}. The three types of driving profiles were considered: safe, moderate, and dangerous. We should note that the profiles (and the model) were not designed with an aim to provide an accurate risk behavior estimation, but rather as a source of plausible ground truth for evaluating SafeML. This dataset was used as the ground truth to train classifiers, whose performance was evaluated through subsequent simulation in CARLA.

For GTSRB, as the dataset is already available, create dataset was assumed to be complete from `Acquisition' phase. Then, a network was built to classify the GTSRB dataset. We built a simple convolutional neural network, as they are known for their superior performance on image applications. We then applied the approach mentioned in \ref{sec:methodology}. This allows to obtain the minimum number of required samples and the distance acceptance threshold for this application. 

\begin{table}[t]
\caption{Properties of Driver Profiles}\label{driverProf}
\centering
\begin{tabular}{|l|l|l|l|}
\hline
\textbf{Property/Driving Profile}                     &  \textbf{Safe}    & \textbf{Moderate} & \textbf{Dangerous} \\
\hline
{\bfseries Max Speed}                 & 30\% below limit  & At limit          & 30\% Above limit\\
{\bfseries Traffic Signs}             & Abide by all      & Ignore 50\%       & Ignore 100\%\\
{\bfseries Automatic Lane-change}     & No                & Yes               & Yes\\
{\bfseries Distance to Other Cars}    & 5 m               & 3 m               & 0 m\\

\hline
\end{tabular}
\end{table}


We trained a CNN network. The network was able to achieve a decent accuracy of around 99.73\%. We remind readers that SafeML is model-agnostic, and other ML models could also have been used. This high accuracy resulted in very few incorrect samples to test SafeML. Thus one of the classes was excluded to be considered as an out-of-scope class which reduced the accuracy to 97.5\%. This added a greater disparity to allow for the validation of SafeML. 


\begin{table}[t]
\caption{Properties of Driver Profiles}\label{modelPerfomance}
\centering
\begin{tabular}{|l|l|l|l|l|}
\hline

\textbf{Model} & \textbf{Class}  &  \textbf{Recall}    & \textbf{Precision}  & \textbf{F1-Score} \\
\hline
kNN & {\bfseries 0}   & 0.89                & 0.95                & 0.92              \\
    &{\bfseries 1}   & 0.0.96            & 0.90                & 0.93              \\
    &{\bfseries 2}   & 0.96                & 0.95                & 0.96              \\
\hline
Random Forest  &{\bfseries 0}   & 0.83                & 0.52                & 0.64  \\
               &{\bfseries 1}   & 0.81                & 0.88                & 0.84  \\
               &{\bfseries 2}   & 0.72                & 0.92                & 0.81    \\
\hline
LSTM           &{\bfseries 0}   & 0.92                & 0.99                & 0.96  \\
               &{\bfseries 1}   & 0.99                & 0.91                & 0.95    \\
               &{\bfseries 2}   & 1.00                & 1.00                & 1.00  \\               
\hline
\end{tabular}
\end{table}

In \cite{Aslansefat2020safeml}, SafeML distance measures have been shown to negatively correlate with the accuracy of the model. From this fact, and according to the first research question established in \Cref{sec:motivation}, we hypothesize that misclassified points would have a higher distance than correctly classified data points due to their dissimilarity to the training set. \\

Furthermore, from principles of statistical analysis, it is established that using an insufficient number of samples during hypothesis testing, there is a risk of the statistical tests not achieving sufficient power. As per our second research question in \Cref{sec:motivation}, our corresponding hypothesis is that the number of samples correlates with confidence of dissimilarity~(the magnitude of the distance). \\

The experiment concludes by following the `Operation' step of the process workflow explained in \Cref{sec:flowchart}. In CARLA example, the same experiment was reproduced in different environment setups to ensure the consistency of the results. In GTSRB, this was performed on the test set, which can be replaced by runtime dataset, at runtime.

\section{Results}
\label{sec:results}

\subsection{Preliminary Findings}
\begin{figure}[t]
\includegraphics[width=\textwidth]{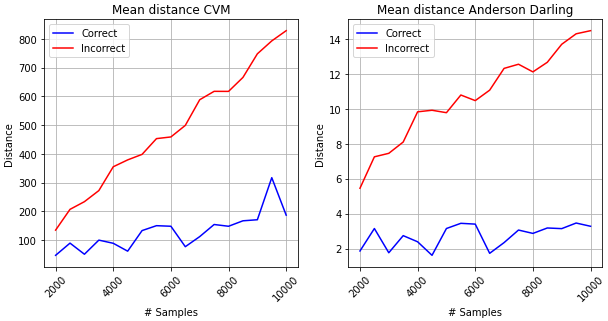}
\caption{Statistical Distance Over Varying Sampling Sizes} \label{fig:sampleSizeVariance}
\end{figure}

Before continuing through the workflow of the simulation, analysis of the trained model was used to test the predefined hypotheses in \Cref{sec:setup}, namely:
\begin{itemize}[leftmargin=25pt]
    \item[\textbf{RQ1}]: \textbf{Dissimilarity-Accuracy Correlation} was tested by calculating the statistical distance between the correctly classified data points and the TSS, as well the incorrectly classified data points and the training scope. \Cref{distanceTable} shows the mean and standard deviation of each of the statistical distance measures used. It shows that the incorrectly classified points are highly dissimilar to the TSS (higher distance), supporting the corresponding hypothesis.
    \item[\textbf{RQ2}]: \textbf{Sample Size Dependence}: Due to the model's accuracy of 80\%, the amount of correctly classified data points were significantly larger than incorrectly classified points when the distances in \Cref{distanceTable} were calculated. To account for the number of samples, the distances were calculated over a varying number of randomly sampled points of each group. As shown in \Cref{fig:sampleSizeVariance}, the distance of incorrectly classified points is always larger than the distance of correctly classified points and increases with increasing number of samples. This can be attributed to a few factors such as: (a) an increased distinction between the distributions and (b) the average value of the distances are shifted with an increased number of available samples, removing skewness in the distribution.
\end{itemize}

\begin{table}[t]
\centering
\caption{Mean and Standard Deviation of Statistical Distances of The Entire Test Set (CVM: Cramer von Mises, AD: Anderson-Darling, KS: Kolmogorov-Smirnov, WS: Wasserstein)}
\label{distanceTable}
\begin{tabular}{|l|l|l|l|l|l|}
\hline

& \textbf{Prediction}     &  \textbf{CVM} & \textbf{AD} & \textbf{KS} & \textbf{WS} \\
\hline

kNN & {\bfseries Correct}    & 1569.71, 617.60  & 8.577, 3.03  & 0.0193, 0.0043 & 3.192e-05, 1.153e-05\\
 & {\bfseries Incorrect}   & 5743.45, 2085.75 & 35.35, 11.12  & 0.083, 0.0139 & 1.430e-04, 5.264e-05 \\

\hline
Random Forest & {\bfseries Correct}     & 3780.74, 227.29  & 18.59, 0.97  & 0.0341, 0.0007 & 1.238e-04, 1.875e-05\\
 & {\bfseries Incorrect}   & 10478.63, 1147.64 & 56.73, 4.78  & 0.1068, 0.0161 & 4.368-04, 6.654e-05 \\
\hline

LSTM & {\bfseries Correct}     & 2744.89, 895.56  & 13.63, 3.26  & 0.0578, 0.0034 & 4.356e-05, 2.276-05\\
 & {\bfseries Incorrect}   & 7892.06, 1033.94 & 43.24, 3.23  & 0.1772, 0.0871 & 2.134e-04, 1.033e-04 \\
\hline

\end{tabular}
\end{table}

\subsection{Experiment Results}
Following the process workflow presented in \Cref{sec:flowchart}, each stage produced its corresponding values after being executed on "Town 1" standard map from CARLA.
In the Acquisition stage, Power Analysis was used on each of the driver profiles. The highest number of samples returned was 91. Multiplying this by an additional factor of 1.3 yields a final number of samples at 120, which aligns with our sampling batches; the operational samples were collected in batches over 4 seconds with a simulation resolution of 30 frames per second.  The performance of the trained model is shown in \Cref{modelPerfomance}, where the kNN model was used in the evaluation of the results due to its simplicity and high reported performance. The resulting threshold values for SafeML are shown in \Cref{thresholdTable}.
    
\begin{table}[t]
\centering
\caption{Threshold Parameters Used for Town 1 (CVM: Cramer von Mises, AD: Anderson-Darling, KS: Kolmogorov-Smirnov, WS: Wasserstein)}\label{thresholdTable}
\begin{tabular}{|l|l|l|l|l|}
\hline

\textbf{Prediction}     &  \textbf{CVM} & \textbf{AD} & \textbf{KS} & \textbf{WS} \\
\hline
{\bfseries Mean}     & 387.83  & 9.64  & 0.087 & 1.38e-4\\
{\bfseries Standard Deviation}   & 171.57 & 3.61  & 0.02 & 6.22e-5 \\

\hline
\end{tabular}
\end{table}

The acceptable performance of the ML-model is a design decision which is obtained from the application requirements specified. In our example, let us consider the correctness over a batch. Since each batch contains multiple frames, let us assume a batch is considered correctly classified if its overall accuracy is 0.8 (96 correct points out of 120). Consequently, a batch is assumed to be incorrectly classified if its overall accuracy is 0 (focusing on worst-case scenarios), where all of its members are misclassified. This high limit was chosen to represent the extreme scenario that minimizes the number of false alarms. 

\begin{figure}[t]
    \centering
    \includegraphics[width=\textwidth]{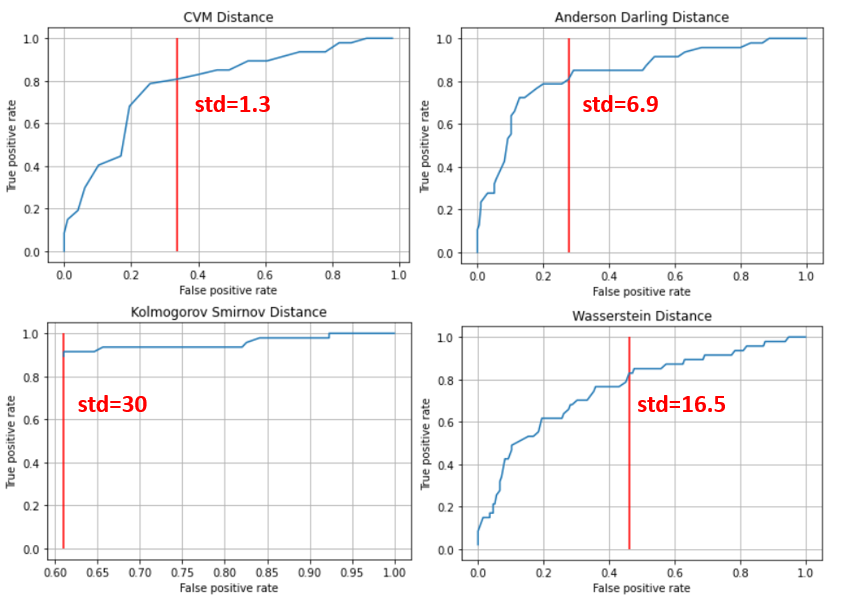}
    \caption{SafeML performance on Town 1 with Moderate Driver Profiles} \label{fig:modDriver}
\end{figure}

\begin{figure}[t]
\includegraphics[width=\textwidth]{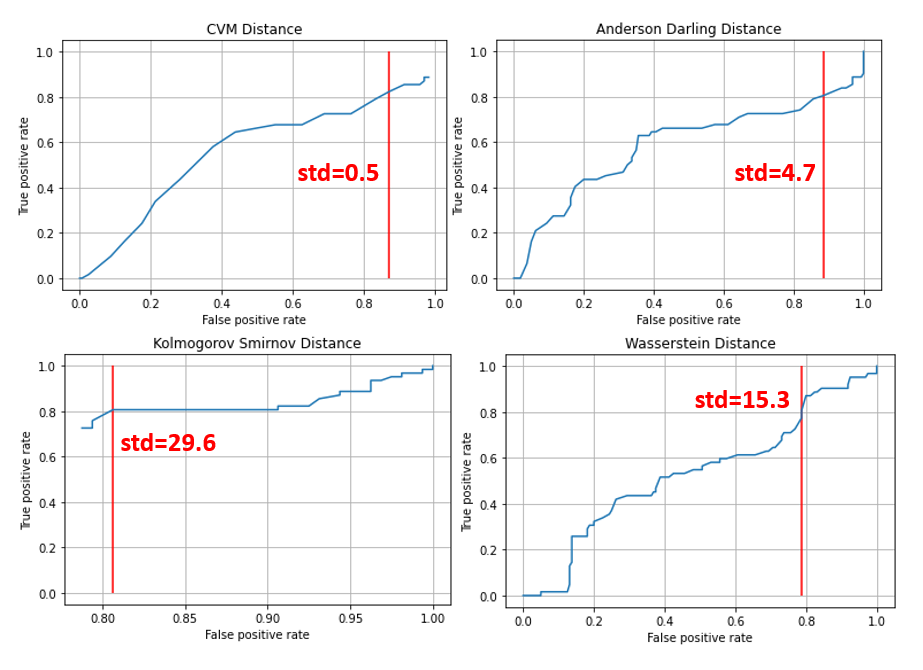}
\caption{SafeML performance on Town 1 With Dangerous Driver Profiles} \label{fig:dangerDriver}
\end{figure}

The performance of each of the distance measures in SafeML was evaluated on different driver profiles as shown in \Cref{fig:modDriver,fig:dangerDriver}, where the true positive rate~(batches with 0 accuracy that were above the threshold) and the false positive rate (batches with 0.8 accuracy that were above the threshold) were plotted over a varying increase in the threshold using steps of 0.1 of the standard deviation.

\Cref{fig:modDriver} shows the factor of standard deviation by which the threshold should be increased to yield reliable identification by SafeML. The plot compares incorrect (i.e. false positive rate) versus correct SafeML alarms (true positive rate), set to a threshold of 0.8 (as mentioned previously, this threshold can be decided based on application-level requirements).
Through this method, a suitable factor for the distance measures was found, with the exception of Kolmogorov Smirnov, where a similar percentage of false positive rate was achieved for the distance measures.

The same process was repeated for the dangerous driver profile in \Cref{fig:dangerDriver}, where similar plot curves were observed, and the threshold points can be established following similar steps as per the moderate profile.
However, the performance ratio between true and false positive rate is exceptionally worse.
The experiment was repeated on "Town 2" and "Town 4" with similar results.

\begin{figure}[t]
\includegraphics[width=\textwidth]{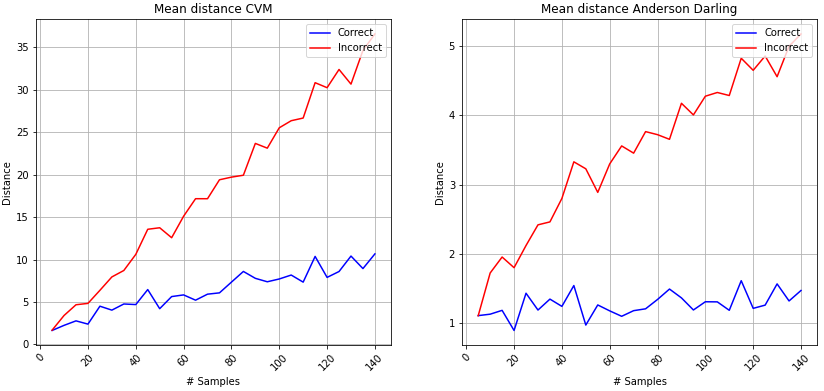}
\caption{Statistical Distance Over Varying Sampling Sizes for GTSRB} \label{fig:GTDSperformance}
\end{figure}

Repeating the process workflow on the GTSRB shows quite a similar trend where the correct and incorrect classification are completely separable by setting a suitable distance threshold as shown in \Cref{fig:GTDSperformance}. The number of samples (where each sample is an image) required can be seen on x-axis. In this case, the majority of incorrect classifications represent an out-of-scope class. The distance was calculated using features derived from the last layer of the CNN, instead of the raw pixels. 

\section{Conclusion and Future Work}
\label{conclusion}

In this paper, we have addressed the challenge of determining sampling and distance thresholds for SafeML, a model-agnostic, assessment tool for scope compliance. Our approach incorporates power sampling during the development stage of the subject ML model, to determine the number of samples necessary to achieve sufficient statistical power while applying the SafeML distance evaluation during the runtime stage. Furthermore, we have proposed means of identifying appropriate distance thresholds, based on the observed performance of the ML model during development-time simulation. We validated our approach experimentally, using a scenario developed in the CARLA automotive simulator as well as the publicly available GTSRB dataset.

Apart from the SafeML applications discussed earlier in \Cref{sec:background}, at the time of writing, additional examples are underway, such as investing SafeML towards cancer detection via x-ray imaging, pedestrian detection, financial investment, and on predictive maintenance.

Regarding future work, we are considering further directions to improve SafeML, including investigating the effect of outlier data, the effect of dataset characteristics (per \cite{oreski2017effects}), using dimensionality reduction, accounting for uncertainty in the dataset labels (per \cite{northcutt2021confidentlearning}), and expanding the scope towards graph, quantum, and time-series datasets.

\section*{Code Availability}
Regarding the research reproducibility, codes and functions supporting this paper are published online at: \url{https://tinyurl.com/4a76z2xs}

\section*{Acknowledgements}
This work was supported by the Secure and Safe Multi-Robot Systems (SES\-AME) H2020 Project under Grant Agreement 101017258 and the German Federal Ministry for Economic Affairs and Climate Action (BMWK) within the research project “FabOS” under grant 01MK20010A.


\bibliographystyle{splncs04}

\end{document}